\pdfoutput=1

\documentclass[11pt]{article}

\usepackage[final]{acl}

\usepackage{times}
\usepackage{latexsym}
\usepackage{todonotes}
\usepackage[T1]{fontenc}
\usepackage{graphicx}
\graphicspath{ {./} }
\usepackage[utf8]{inputenc}

\usepackage{microtype}
\usepackage{caption}
\usepackage{subcaption}
\usepackage{pgfplots}
\pgfplotsset{width=8cm,height=6cm,compat=1.9}

\title{Scalable Training of Language Models using JAX pjit and TPUv4}

\author{
\begin{tabular}{c@{\extracolsep{2em}}c@{\extracolsep{2em}}c@{\extracolsep{2em}}c}
Joanna Yoo\thanks{* These authors contributed equally.}   & Kuba Perlin\footnotemark[1] & Siddhartha Rao Kamalakara & João G.M. Araújo \\
\end{tabular} \\
Cohere \\
\texttt{\{joanna,kuba,sid,joaogui1\}@cohere.ai}
}

\usepackage{xspace}
\newcommand{\pjit}{\texttt{pjit}\xspace}

\begin{document}
\maketitle
\begin{abstract}
Modern large language models require distributed training strategies due to their size. The challenges of efficiently and robustly training them are met with rapid developments on both software and hardware frontiers. In this technical report, we explore challenges and design decisions associated with developing a scalable training framework, and present a quantitative analysis of efficiency improvements coming from adopting new software and hardware solutions.
\end{abstract}

\section{Introduction}

Scaling up is one of the most common ways of obtaining better language models~\citep{T5}. Once the model size becomes large enough to prohibit fitting the entire model on a single device, new challenges arise. On the hardware side, extensive amounts of compute resources with large memory and fast interconnect are needed. On the software side, algorithms need to be developed that efficiently utilize that hardware, and optimize the time and resources necessary to train a model.

This technical report explores the challenges our team has faced when scaling language models to hundreds of billions of parameters, and how our proprietary framework, FAX, is designed to address those challenges. We focus on the breakthroughs in training efficiency achieved by using JAX~\citep{jax} to enable tensor and data parallelism through GSPMD~\citep{gspmd}, XLA, and Google Cloud TPU VMs~\citep{tpu-vm}, highlighting the use of recently released TPU~v4 Pods~\citep{tpu-v4}.

\section{The FAX Framework} 


To accelerate research, development, and production of large language models, the underlying training framework should make it easy to experiment with new model architectures and training algorithms, while also being seamlessly scalable and easy to integrate with other workflows, such as model evaluation or data processing. 


That motivates the modular design of FAX, which allows for a maintainable and malleable codebase, in which the data loading, hardware acquisition, model serialization, model definition, and compilation modules are all logically separate, making it easy to change any one of those parts in isolation, as new advancements are made.


\subsection{Model Definition}
\begin{figure}
    \centering
    \includegraphics[width=0.48\textwidth]{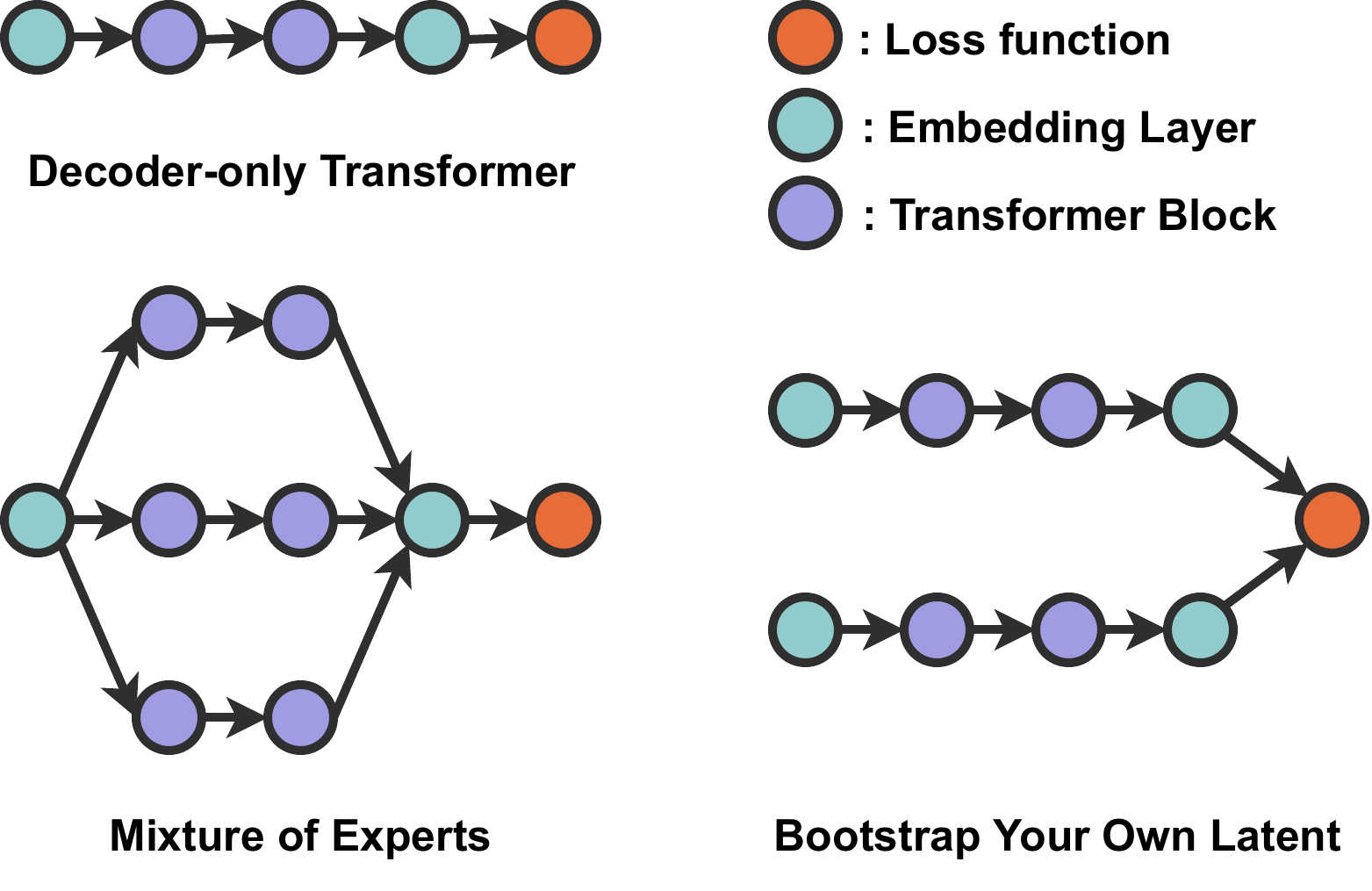}
    \caption{Simplified DAG representations of a variety of model architectures: decoder-only Transformer~\citep{transformer}, Mixture of Experts ~\citep{moe}, and Bootstrap Your Own Latent~\citep{byol}. 
    A versatile training framework should make a wide range of architectures, including those, easy to implement.
    }
    \label{fig:dags}
\end{figure}

A user's configuration file is parsed into a precise abstract description of a machine learning model. The description is constrained to the form of a Directed Acyclic Graph (DAG), the nodes of which are parameterized functions (see Fig.~\ref{fig:dags}). Those often correspond to entities usually referred to as `layers', such as Transformer blocks, but developers are free to define any nodes they wish.

Having this abstract intermediate representation of a model opens up possibilities for static analysis -- for example, the memory requirements or latency estimates for the model can be computed at this point. Such analysis can be used to aid model compilation and experiment design.

\subsection{Model Compilation using pjit}

Given the abstract model definition and available hardware resources, this module compiles the model's training step (and other required functions, e.g. validation step) into a function executed on all the provided hardware in a distributed fashion. 

Multiple implementations of this step can be written, and switched between, thanks to the modularity of the framework. For example, the entire model may be compiled, with tensor-parallelism, onto a single hardware unit (such as a GPU, or a TPU Pod slice), or a strategy may be devised for partitioning the model DAG over multiple heterogeneous hardware units.

The backbone of FAX model compilation is the \pjit (partitioned just-in-time compilation) feature of JAX, which allows for compiling an arbitrary JAX function into an SPMD (single program, multiple data) XLA computation that runs on multiple devices, potentially on multiple hosts. For each \pjit-ted function, the programmer only needs to specify how the inputs and outputs of the function shall be partitioned, although one can also add custom sharding constraints for any intermediate variables inside the function to further control the compilation. 

The use of \pjit requires specifying a logical mesh of devices, which is simply an $n$-dimensional array of physical devices (e.g. TPU cores). Each of the dimensions of the logical mesh may be referred to as a `logical mesh axis' and has an associated name. The partitioning specification for inputs and outputs of \pjit-ted functions is then a description of which axes of the tensors should be partitioned across which logical mesh axes. A single axis of a tensor may be partitioned across one, none, or multiple logical mesh axes. For example, one may use a $2$-dimensional logical mesh with `data parallelism' and `tensor parallelism' axes, and then partition the batch and embedding dimensions of all activations along those two logical axes, respectively.

\pjit makes it straightforward to implement data and tensor parallelism, and experiment with the trade-offs between the two, and is a key mechanism powering the FAX training framework.

\section{Model Parallelism in FAX}\label{sec:model-parallelism} 

\begin{figure}[t]
    \centering
    \includegraphics[width=0.48\textwidth]{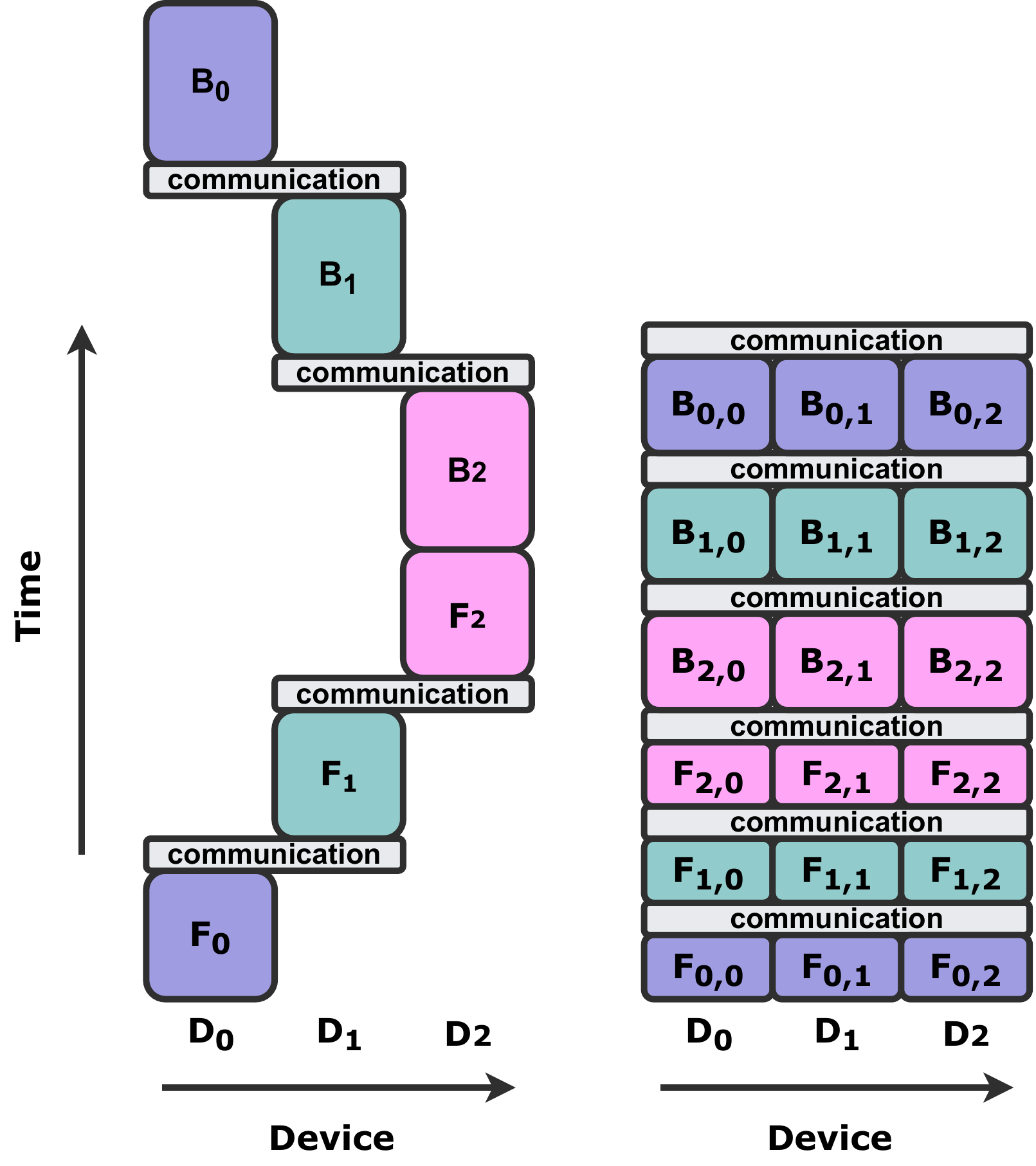}
    \caption{A simplified timing diagram (not to scale) of a forward and backward pass through three layers, distributed across three devices. Left: pipeline parallelism, right: tensor parallelism. $F_i$ and $B_i$ denote the forward and backward pass of the $i$th layer, respectively. In the tensor parallelism paradigm, each of those computations is further partitioned into smaller computations $F_{i,j}$ and $B_{i,j}$, executed on devices $D_j$. The idleness in pipeline parallelism can be traded off for increased inter-device communication in tensor parallelism.}
    \label{fig:pp-vs-tp}
\end{figure}

Contemporary large language models are too big to be trained on a single accelerator.
A $6.7$B-parameter model, for example, requires $25$GiB of memory for just the parameters (stored as float32). Even a single state-of-the-art GPU (e.g. a $40$GB NVIDIA A100) or a TPUv4 accelerator may not be large enough to train such a model, once memory required for the optimizer state is accounted for. Thus, partitioning the model across multiple accelerators is necessary. Two common ways of doing that are \textit{pipeline parallelism} and \textit{tensor parallelism}, illustrated in Fig.~\ref{fig:pp-vs-tp}.

\subsection{Pipeline Parallelism}\label{ssec:pp}

A pipeline parallelism solution, such as GPipe~\citep{gpipe}, would partition the model into groups of consecutive layers. Each batch of data then needs to be processed on one accelerator, then the following one, and so on. After the forward pass, the backward pass needs to be run on the accelerators in reversed order. 

Pipeline parallelism is particularly effective when the user has access to multiple, possibly heterogeneous, compute units without fast interconnect between them. The forward and backward functions of the given layers can be compiled independently, supporting heterogeneity, and the only communication is the passing of activations and gradients once per device, as opposed to tensor parallelism, which requires collective communication after most of model layers.

The main drawback of pipeline parallelism is that the accelerators spend significant amounts of time idle. Methods of decreasing the idleness exist, but come at a cost; for example, PipeDream~\citep{pipedream} achieves that by introducing parameter staleness.

\subsection{Tensor Parallelism}\label{ssec:tp}

An alternative form of model parallelism, which introduces different trade-offs, is \textit{tensor parallelism} -- partitioning large tensors (model weights and activations alike) across accelerators. Large computations, such as matrix multiplications, can then be performed in parallel across multiple devices and collated. The frequent need for all-gather operations may seem worrying, but with fast enough interconnect (such as that in TPU Pods), tensor parallelism can be the most efficient solution.

We apply tensor parallelism to our transformer models by sharding all the large model weight tensors and the activations. This places almost no constraints on the model, albeit we do require that the number of attention heads be divisible by the order of tensor parallelism, so that the heads of multi-headed attention blocks can be distributed across accelerators. \pjit correctly infers the optimal sharding for any intermediate tensors, wherever we do not specify sharding constraints explicitly. 

\subsection{Combining Multiple Kinds of Parallelism}
The model parallelism methods mentioned above are not mutually exclusive - in fact, both types of parallelism can be combined, and furthermore, they can be combined with \textit{data parallelism}, which means processing different subsets of a batch of data on different devices in parallel.

We found that as long as the model resides on a single hardware unit that has fast interconnect between the accelerators, it is sufficient and even optimal to use tensor and data parallelism only. We present quantitative results, supporting that statement, in Section~\ref{ssec:pp_vs_tp}.

\section{TPU~v4 and Multi-Host Training}\label{sec:hardware}

Two hardware types most commonly used for training large neural networks, at present, are Graphics Processing Units (GPUs) and Google's Tensor Processing Units (TPUs)~\citep{gpu-tpu-benchmarks}. 
We have primarily been using the latter for our large scale training.

\subsection{Using TPU VMs for Distributed Training}

Since 2021, Google Cloud offers direct access to TPU VMs -- a user can connect directly to a TPU host machine, with direct access to a number of TPU chips, set up their environment there, and run code directly on the local devices. That procedure can be scaled up to a larger TPU Pod slice (set of TPU hosts residing in a single physical Pod) containing multiple TPU VMs. Each VM may run the same code on different data, and the VMs may communicate over the Pod's fast interconnect.

Running JAX on TPU Pods requires users to run the code on all hosts in the Pod simultaneously. The users need to explicitly handle coordination of execution across TPU VMs. We use Ray~\citep{ray}, a distributed computing framework, to achieve that. FAX establishes a Ray node cluster consisting of a main host VM and TPU VMs. The host is responsible for sending code and artifacts required for training to the TPU VMs, running the training step on each TPU VM using remote calls, and retrieving metrics and other outputs.

\subsection{Google Cloud TPU~v4 Pods}\label{ssec:tpu}

We have recently been granted access to Google's new $4$th generation TPUs, which more than double the computational power of their predecessor, TPU~v3. TPU~v4 cores provide $275$ peak TFLOPS of compute power (compared to TPU~v3's $122$ peak TFLOPS).
That increase in performance has enabled us to iterate on ideas and validate them at a much faster pace than before.

\section{Results}
\subsection{Comparison to Previous Framework}\label{ssec:pp_vs_tp}

\begin{figure}[t]
    \centering
    \includegraphics[width=0.48\textwidth]{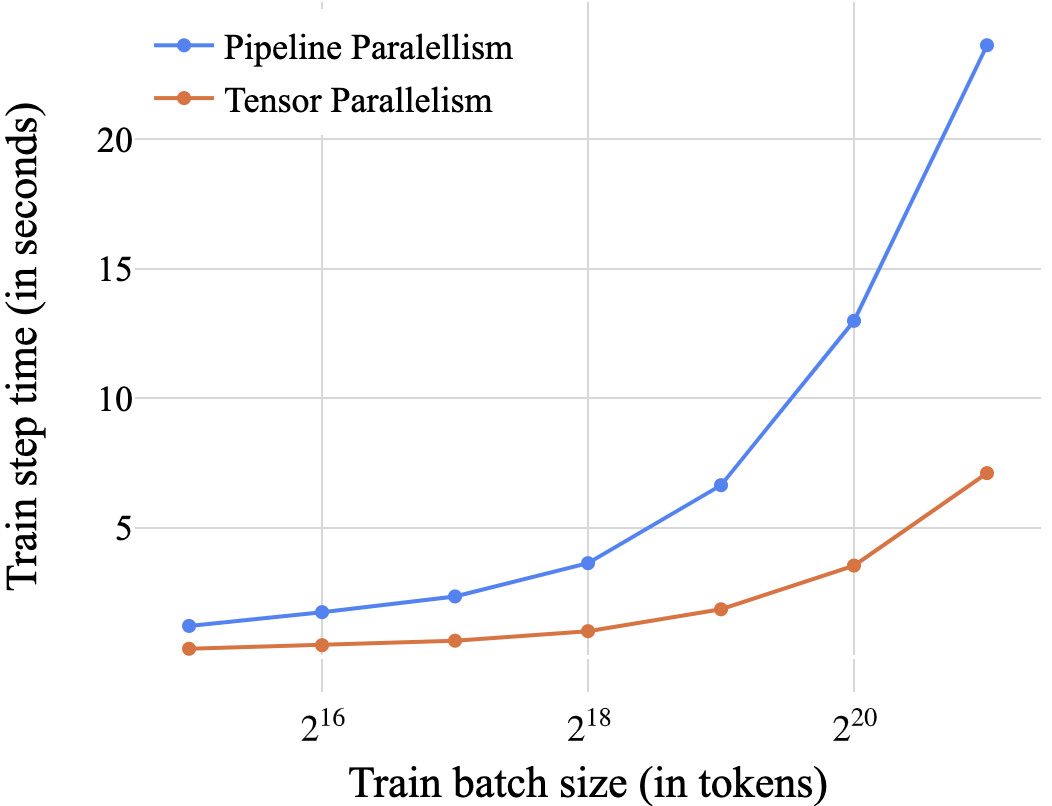}
    \caption{Comparison of the training step time of a $350$M-parameter transformer model on TPU~v4-32, using different kinds of model parallelism. The pipeline-parallel training is optimized with GPipe~\citep{gpipe}, with optimal micro-batch size, while the tensor-parallel training uses the optimal sharding configuration.}
    \label{fig:wit-vs-fax}
\end{figure}

Our previous proprietary training framework is built with TensorFlow~\citep{tensorflow} and uses pipeline parallelism as the only means of scaling the model. The unsuitability of that method for our setting, related to sub-optimal hardware utilization (`pipeline bubbles'), is illustrated in Fig.~\ref{fig:wit-vs-fax}.

\subsection{Model Scalability with TPU~v4}\label{ssec:v3_vs_v4}

While FAX supports pipeline parallelism, given the number of TPU~v4 cores available to us, we are able to train extremely large models using tensor and data parallelism only. The capacity of different sizes of TPU~v4 Pod slices is presented in Table~\ref{tab:v4-tpu-vm-capacities}.

\begin{table}
    \centering
    \begin{tabular}{|c|c|c|}\hline
        Pod slice & max model size & step time \\\hline
        v4-16 & $13.7$B & $0.87$s$\pm0.02$s \\ 
        v4-128 & $86.6$B & $1.52$s$\pm0.02$s \\ 
        v4-512 & $340.0$B & $6.21$s$\pm0.08$s\\ 
        \hline 
    \end{tabular}
    \caption{Sizes of transformer models that can be trained on a single TPU~v4 Pod slice, without the need for pipeline parallelism. The figures assume a batch size of $1$, sequence length of 2048, and maximal tensor parallelism. The Adam optimizer~\citep{adam} was used for all experiments. The hidden dimensions were fixed at $5120, 10240, 16384$, for the three Pod slice sizes respectively, while the number of layers was maximized, up to a precision of $10$ layers. Standard deviation of step time over $10$ steps is given.
    }
    \label{tab:v4-tpu-vm-capacities}
\end{table}

\begin{figure}
    \centering
    \includegraphics[width=0.50\textwidth]{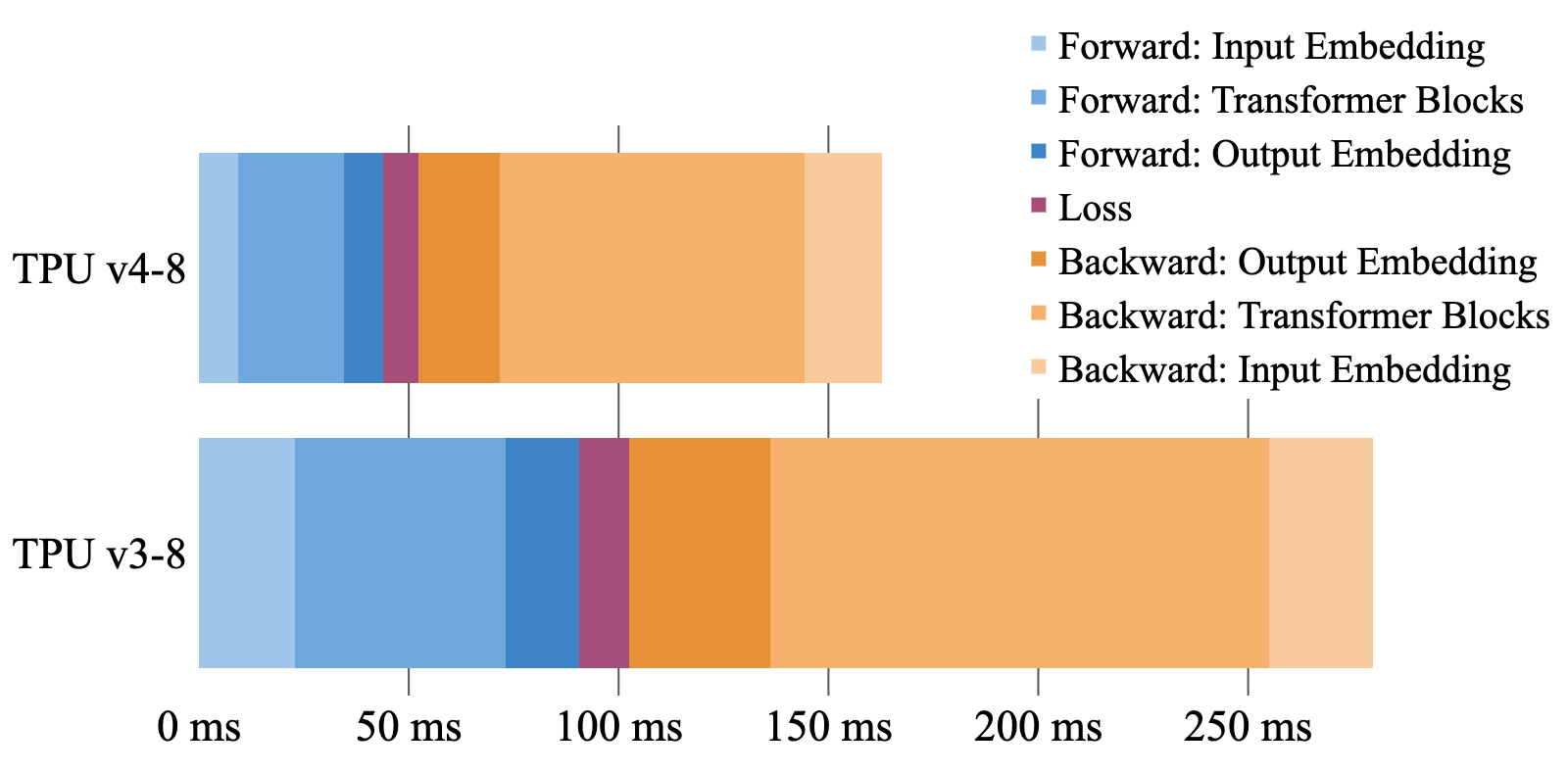}
    \caption{Time spent computing the forward pass, loss, and backward pass of a $124$M-parameter transformer model (batch size $32$, sequence length $2048$). The superior compute power of TPU~v4 chips is reflected in the observed $1.7$x speedup of the total step time. The speedups for the forward and backward passes are $2$x and $1.6$x, respectively, suggesting our backward pass spends a larger portion of time on communication.}
    \label{fig:comp-vs-comm}
\end{figure}

Transitioning from TPU~v3 to TPU~v4, we have so far achieved a total speedup of around $1.7$x (see Fig.~\ref{fig:comp-vs-comm}). 
It is a remaining challenge to optimize our model compilation to maximally benefit from the faster computation of TPU~v4.

\section{Challenges}

The possibilities unlocked by distributed training come at a cost of having to manage the complexity of distributed systems. Failures of individual hardware hosts do occur, and minimizing the lost computation and engineering effort necessary to restart training is a complex problem.

A further lesson we have learned is that ensuring correctness of a framework for training large neural networks is difficult, because the networks tend to still achieve good (but sub-optimal) performance in the presence of non-breaking bugs, and problems that result in a slight performance deterioration may long stay undetected.

\section{Conclusions}  


The design of a versatile distributed training framework needs to address multiple challenges. Meticulous codebase design is required to enable seamless scalability, efficient utilization of powerful compute, and rapid prototyping of models and training algorithms.

We have described the modular design of our training framework and its utilization of state-of-the-art hardware (TPU~v4) and software (JAX and pjit) to perform efficient, large-scale, parallel computation. Our experiments illustrate the capabilities of TPU~v4, and the importance of the choice of training parallelization strategy.


\clearpage
\section*{Acknowledgements}

We thank James Bradbury, Skye Wanderman-Milne, Vaibhav Singh, Allen Wang, Yash Katariya, Almond Au, and other members of the Google JAX and TPU teams, for their support in working with their technologies.

\bibliography{anthology,custom}

\begin{thebibliography}{14}
\expandafter\ifx\csname natexlab\endcsname\relax\def\natexlab#1{#1}\fi

\bibitem[{Abadi et~al.(2015)Abadi, Agarwal, Barham, Brevdo, Chen, Citro,
  Corrado, Davis, Dean, Devin, Ghemawat, Goodfellow, Harp, Irving, Isard, Jia,
  Jozefowicz, Kaiser, Kudlur, Levenberg, Man\'{e}, Monga, Moore, Murray, Olah,
  Schuster, Shlens, Steiner, Sutskever, Talwar, Tucker, Vanhoucke, Vasudevan,
  Vi\'{e}gas, Vinyals, Warden, Wattenberg, Wicke, Yu, and Zheng}]{tensorflow}
Mart\'{\i}n Abadi, Ashish Agarwal, Paul Barham, Eugene Brevdo, Zhifeng Chen,
  Craig Citro, Greg~S. Corrado, Andy Davis, Jeffrey Dean, Matthieu Devin,
  Sanjay Ghemawat, Ian Goodfellow, Andrew Harp, Geoffrey Irving, Michael Isard,
  Yangqing Jia, Rafal Jozefowicz, Lukasz Kaiser, Manjunath Kudlur, Josh
  Levenberg, Dan Man\'{e}, Rajat Monga, Sherry Moore, Derek Murray, Chris Olah,
  Mike Schuster, Jonathon Shlens, Benoit Steiner, Ilya Sutskever, Kunal Talwar,
  Paul Tucker, Vincent Vanhoucke, Vijay Vasudevan, Fernanda Vi\'{e}gas, Oriol
  Vinyals, Pete Warden, Martin Wattenberg, Martin Wicke, Yuan Yu, and Xiaoqiang
  Zheng. 2015.
\newblock \href {http://tensorflow.org/} {{TensorFlow}: Large-scale machine
  learning on heterogeneous systems}.
\newblock Software available from tensorflow.org.

\bibitem[{Bradbury et~al.(2018)Bradbury, Frostig, Hawkins, Johnson, Leary,
  Maclaurin, Necula, Paszke, Vander{P}las, Wanderman-{M}ilne, and Zhang}]{jax}
James Bradbury, Roy Frostig, Peter Hawkins, Matthew~James Johnson, Chris Leary,
  Dougal Maclaurin, George Necula, Adam Paszke, Jake Vander{P}las, Skye
  Wanderman-{M}ilne, and Qiao Zhang. 2018.
\newblock \href {http://github.com/google/jax} {{JAX}: composable
  transformations of {P}ython+{N}um{P}y programs}.

\bibitem[{Grill et~al.(2020)Grill, Strub, Altch\'{e}, Tallec, Richemond,
  Buchatskaya, Doersch, Avila~Pires, Guo, Gheshlaghi~Azar, Piot, kavukcuoglu,
  Munos, and Valko}]{byol}
Jean-Bastien Grill, Florian Strub, Florent Altch\'{e}, Corentin Tallec, Pierre
  Richemond, Elena Buchatskaya, Carl Doersch, Bernardo Avila~Pires, Zhaohan
  Guo, Mohammad Gheshlaghi~Azar, Bilal Piot, koray kavukcuoglu, Remi Munos, and
  Michal Valko. 2020.
\newblock \href
  {https://proceedings.neurips.cc/paper/2020/file/f3ada80d5c4ee70142b17b8192b2958e-Paper.pdf}
  {Bootstrap your own latent - a new approach to self-supervised learning}.
\newblock In \emph{Advances in Neural Information Processing Systems},
  volume~33, pages 21271--21284. Curran Associates, Inc.

\bibitem[{Huang et~al.(2019)Huang, Cheng, Bapna, Firat, Chen, Chen, Lee, Ngiam,
  Le, Wu et~al.}]{gpipe}
Yanping Huang, Youlong Cheng, Ankur Bapna, Orhan Firat, Dehao Chen, Mia Chen,
  HyoukJoong Lee, Jiquan Ngiam, Quoc~V Le, Yonghui Wu, et~al. 2019.
\newblock Gpipe: Efficient training of giant neural networks using pipeline
  parallelism.
\newblock \emph{Advances in neural information processing systems},
  32:103--112.

\bibitem[{Kingma and Ba(2015)}]{adam}
Diederik~P. Kingma and Jimmy Ba. 2015.
\newblock \href {http://arxiv.org/abs/1412.6980} {Adam: {A} method for
  stochastic optimization}.
\newblock In \emph{3rd International Conference on Learning Representations,
  {ICLR} 2015, San Diego, CA, USA, May 7-9, 2015, Conference Track
  Proceedings}.

\bibitem[{Moritz et~al.(2018)Moritz, Nishihara, Wang, Tumanov, Liaw, Liang,
  Elibol, Yang, Paul, Jordan, and Stoica}]{ray}
Philipp Moritz, Robert Nishihara, Stephanie Wang, Alexey Tumanov, Richard Liaw,
  Eric Liang, Melih Elibol, Zongheng Yang, William Paul, Michael~I. Jordan, and
  Ion Stoica. 2018.
\newblock \href {http://arxiv.org/abs/1712.05889} {Ray: A distributed framework
  for emerging ai applications}.

\bibitem[{Narayanan et~al.(2019)Narayanan, Harlap, Phanishayee, Seshadri,
  Devanur, Ganger, Gibbons, and Zaharia}]{pipedream}
Deepak Narayanan, Aaron Harlap, Amar Phanishayee, Vivek Seshadri, Nikhil~R.
  Devanur, Gregory~R. Ganger, Phillip~B. Gibbons, and Matei Zaharia. 2019.
\newblock \href {https://doi.org/10.1145/3341301.3359646} {Pipedream:
  Generalized pipeline parallelism for dnn training}.
\newblock In \emph{Proceedings of the 27th ACM Symposium on Operating Systems
  Principles}, SOSP '19, page 1–15, New York, NY, USA. Association for
  Computing Machinery.

\bibitem[{Raffel et~al.(2020)Raffel, Shazeer, Roberts, Lee, Narang, Matena,
  Zhou, Li, and Liu}]{T5}
Colin Raffel, Noam Shazeer, Adam Roberts, Katherine Lee, Sharan Narang, Michael
  Matena, Yanqi Zhou, Wei Li, and Peter~J. Liu. 2020.
\newblock \href {http://jmlr.org/papers/v21/20-074.html} {Exploring the limits
  of transfer learning with a unified text-to-text transformer}.
\newblock \emph{J. Mach. Learn. Res.}, 21:140:1--140:67.

\bibitem[{Selvan and Kanwar(2021)}]{tpu-v4}
Aarush Selvan and Pankaj Kanwar. 2021.
\newblock Google showcases cloud tpu v4 pods for large model training.
\newblock
  \url{https://cloud.google.com/blog/topics/tpus/google-showcases-cloud-tpu-v4-pods-for-large-model-training}.
\newblock Accessed: 2022-02-21.

\bibitem[{Shazeer et~al.(2017)Shazeer, Mirhoseini, Maziarz, Davis, Le, Hinton,
  and Dean}]{moe}
Noam Shazeer, Azalia Mirhoseini, Krzysztof Maziarz, Andy Davis, Quoc~V. Le,
  Geoffrey~E. Hinton, and Jeff Dean. 2017.
\newblock \href {https://openreview.net/forum?id=B1ckMDqlg} {Outrageously large
  neural networks: The sparsely-gated mixture-of-experts layer}.
\newblock In \emph{5th International Conference on Learning Representations,
  {ICLR} 2017, Toulon, France, April 24-26, 2017, Conference Track
  Proceedings}. OpenReview.net.

\bibitem[{Spiridonov(2021)}]{tpu-vm}
Alexander Spiridonov. 2021.
\newblock New cloud tpu vms make training your ml models on tpus easier than
  ever.
\newblock
  \url{https://cloud.google.com/blog/products/compute/introducing-cloud-tpu-vms}.
\newblock Accessed: 2022-02-21.

\bibitem[{Vaswani et~al.(2017)Vaswani, Shazeer, Parmar, Uszkoreit, Jones,
  Gomez, Kaiser, and Polosukhin}]{transformer}
Ashish Vaswani, Noam Shazeer, Niki Parmar, Jakob Uszkoreit, Llion Jones,
  Aidan~N Gomez, Lukasz Kaiser, and Illia Polosukhin. 2017.
\newblock Attention is all you need.
\newblock \emph{arXiv preprint arXiv:1706.03762}.

\bibitem[{Wang et~al.(2019)Wang, Wei, and Brooks}]{gpu-tpu-benchmarks}
Yu~Wang, Gu{-}Yeon Wei, and David Brooks. 2019.
\newblock \href {http://arxiv.org/abs/1907.10701} {Benchmarking tpu, gpu, and
  {CPU} platforms for deep learning}.
\newblock \emph{CoRR}, abs/1907.10701.

\bibitem[{Xu et~al.(2021)Xu, Lee, Chen, Hechtman, Huang, Joshi, Krikun,
  Lepikhin, Ly, Maggioni, Pang, Shazeer, Wang, Wang, Wu, and Chen}]{gspmd}
Yuanzhong Xu, HyoukJoong Lee, Dehao Chen, Blake~A. Hechtman, Yanping Huang,
  Rahul Joshi, Maxim Krikun, Dmitry Lepikhin, Andy Ly, Marcello Maggioni,
  Ruoming Pang, Noam Shazeer, Shibo Wang, Tao Wang, Yonghui Wu, and Zhifeng
  Chen. 2021.
\newblock \href {http://arxiv.org/abs/2105.04663} {{GSPMD:} general and
  scalable parallelization for {ML} computation graphs}.
\newblock \emph{CoRR}, abs/2105.04663.

\end{thebibliography}
\bibliographystyle{acl_natbib}

\end{document}